\pdfoutput=1

\documentclass[11pt]{article}

\usepackage{EMNLP2022}

\usepackage{times}
\usepackage{latexsym}

\usepackage[T1]{fontenc}

\usepackage[utf8]{inputenc}

\usepackage{microtype}

\usepackage{inconsolata}

\usepackage{graphicx}
\usepackage{xcolor}
\usepackage{arabtex}
\usepackage{utf8}
\usepackage[utf8]{inputenc}
\setcode{utf8}

\usepackage{comment}
\usepackage{todonotes}
\usepackage{color,soul}

\newcommand{\squeezeupless}{\vspace{-2mm}}
\newcommand\blfootnote[1]{%
  \begingroup
  \renewcommand\thefootnote{}\footnote{#1}%
  \addtocounter{footnote}{-1}%
  \endgroup
}

\title{NatiQ: An End-to-end Text-to-Speech System for Arabic}

\author{Ahmed Abdelali\textsuperscript{1}, Nadir Durrani\textsuperscript{1}, Cenk Demiroglu\textsuperscript{2}, \\\textbf{Fahim Dalvi\textsuperscript{1}, Hamdy Mubarak\textsuperscript{1}, Kareem Darwish\textsuperscript{3}} \\
  \textsuperscript{1}Qatar Computing Research Institute - Hamad Bin Khalifa University, Doha, Qatar\\
\textsuperscript{2}Özyeğin University, Istanbul, Türkiye \textsuperscript{3}aiXplain Inc. Los Gatos, CA, USA\\
  \textsuperscript{1}\texttt{{aabdelali,ndurrani,fdalvi,hmubarak}@hbku.edu.qa}\\
  \textsuperscript{2}\texttt{cenk.demiroglu@ozyegin.edu.tr} \textsuperscript{3}\texttt{kareem.darwish@aixplain.com}\\}

\begin{document}
\maketitle
\begin{abstract}
\textbf{NatiQ} is end-to-end text-to-speech system for Arabic. 
Our speech synthesizer uses an encoder-decoder architecture with attention. 
 We used both tacotron-based models (tacotron-1 and tacotron-2) and the faster transformer model for generating mel-spectrograms from characters. 
 We concatenated Tacotron1 with the WaveRNN vocoder, Tacotron2 with the WaveGlow  vocoder and ESPnet transformer with the parallel wavegan vocoder to synthesize  waveforms from the spectrograms. We used 
in-house speech data for two voices: 1) neutral male ``Hamza''- narrating general content and news, and 2) expressive female ``Amina''- narrating children story books to train our models. Our best systems 
achieve an average Mean Opinion Score (MOS) of 4.21  and 4.40 for Amina and Hamza respectively.
The objective evaluation of the systems using word and character error rate (WER and CER) as well as the response time measured by real-time factor favored the end-to-end architecture ESPnet.
NatiQ demo is available online at\\ \url{https://tts.qcri.org}. 
\end{list}
\end{abstract}

\section{Introduction}
\label{sec:intro}
\blfootnote{This works was done while Kareem Darwish was at Qatar Computing Research Institute.}

Text to speech (TTS) is among the technologies that 
enables many solutions 
across different sectors. 
In the current pandemic time, education system is challenged with the new norm of distance and remote education. Teachers are not able to provide needed attention and support for every student; more precisely for lower elementary schools where students are very dependent on the teacher's guidance to follow the instructions. 
TTS can elevate some of this burden by allowing the young children to hear the content and have it read to them in a very fluent and pleasing voice. Advances in Neural technology allow achieving 
more natural 
voice compared to previous technologies~\cite{arxiv.1905.00590}. 

\begin{figure}[htb!]
   \centering
     \includegraphics[width=0.36\textwidth]{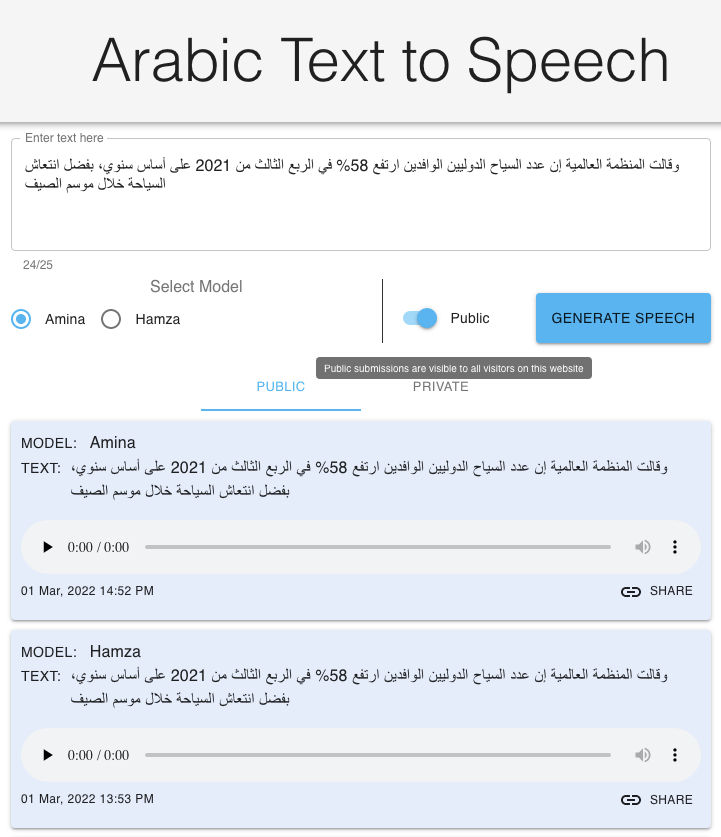}
     \caption{NatiQ system in action 
     }
     \label{fig:natiq}
     \vspace{-1mm}
 \end{figure}

We present \textbf{NatiQ}, an end-to-end speech system for Arabic. The system is 
composed of two independent modules: i) the web application and ii) the speech synthesizer. The web application uses \emph{React Javascript} framework to handle dynamic User Interface and \emph{MangoDB} to handle session related information. The system is built upon modern web technologies, allowing it to run 
cross-browsers and platforms.
Figure \ref{fig:natiq} presents a screenshot of the interface. 

Our best synthesizer is based on ESPnet Transformer TTS~\cite{10.1609/aaai.v33i01.33016706}
architecture that takes input characters in an encoder-decoder framework to output mel-spectograms. 
The intermediate form is 
then converted into 
wav form using the Generative Adversarial Networks vocoder WaveGAN~\cite{Adversarialaudio2019Donahue,Yamamoto2020ParallelWA}. We explored additional architecture including Tacotron1~\cite{wang2017tacotron} and 2~\cite{shen2018natural} and for vocoders  WaveRNN \cite{kalchbrenner2018efficient} and WaveGlow \cite{prenger2018waveglow} to synthesize waveforms from the decoded mel-spectograms. 

We built two in-house speech corpora 
\emph{Amina} 
-- a female speaker with expressive narration and \emph{Hamza} -- a male speaker with neutral narration. 
The former is targeted towards education and the latter is more suitable to broadcast media. 

Given that Arabic is typically written with no short vowels, this required to include additional processing to the text before exploiting it in the training. In addition to the short vowels restoration, diacritization, the pre-processing steps involves segmentation, transcript matching, voice normalization and silence reduction. We will further describe the pipeline and the architecture in detail.
The resulting systems were evaluated using both objective and subjective approaches employing automatic metrics such as CER and WER; and using MOS. Lastely, the  systems were assessed with Real-time Factor to evaluated decoding speed of each model.



\begin{figure}[htb!]
   \centering
     \includegraphics[width=0.40\textwidth]{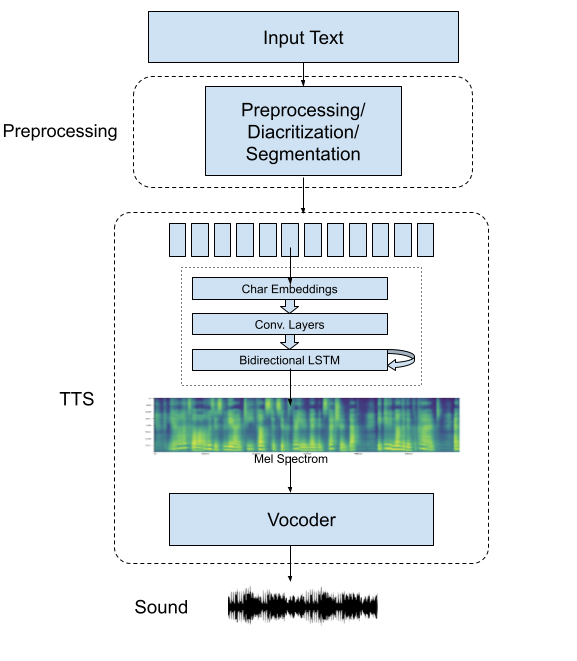}
     \caption{NatiQ Architecture. 
     }
     \label{fig:natiqArch}
     \vspace{-1mm}
 \end{figure} 
 
\section{System Architecture}
Our NatiQ system is a web-based demonstration that is composed of two main components: 

\subsection{Web Application}

The web application has two major components; the frontend and the backend. The frontend is created using the React Javascript framework to handle the dynamic User Interface (UI) changes such updates in generation. The backend is built using NodeJS and MongoDB to handle sessions, data associated with these sessions, communication with models, request inference and authentication. The frontend presents the user with an input text box and choice of speakers to choose from. Figure~\ref{fig:natiq} shows a screenshot for the frontend. The responses from the backend will be presented to the user in a wave form that the user can listen to or download.  

\subsection{Speech Synthesis}

Now we will describe the overall architecture of our synthesis model. Figure \ref{fig:natiqArch} 
shows the system architecture. The preprocessing module involves converting the numbers, abbreviations and dates into their vocalized form using linguistic and custom rules. Next the text is vowelized using Farasa \cite{abdelali-etal-2016-farasa}, which diacritize and restore short vowels using the syntactic structure of the sentence. 

The synthesizer is an encoder-decoder model cascaded with a vocoder to generate the wave-forms. The 
former 
converts the preprocessed text into 
a mel-spectrum. The 
latter convert the 
melspectogram representation into a wave form. 
Below we describe different components of our model:



\subsubsection{Data}

\begin{figure}[htb!]
   \centering
     \includegraphics[width=0.35\textwidth]{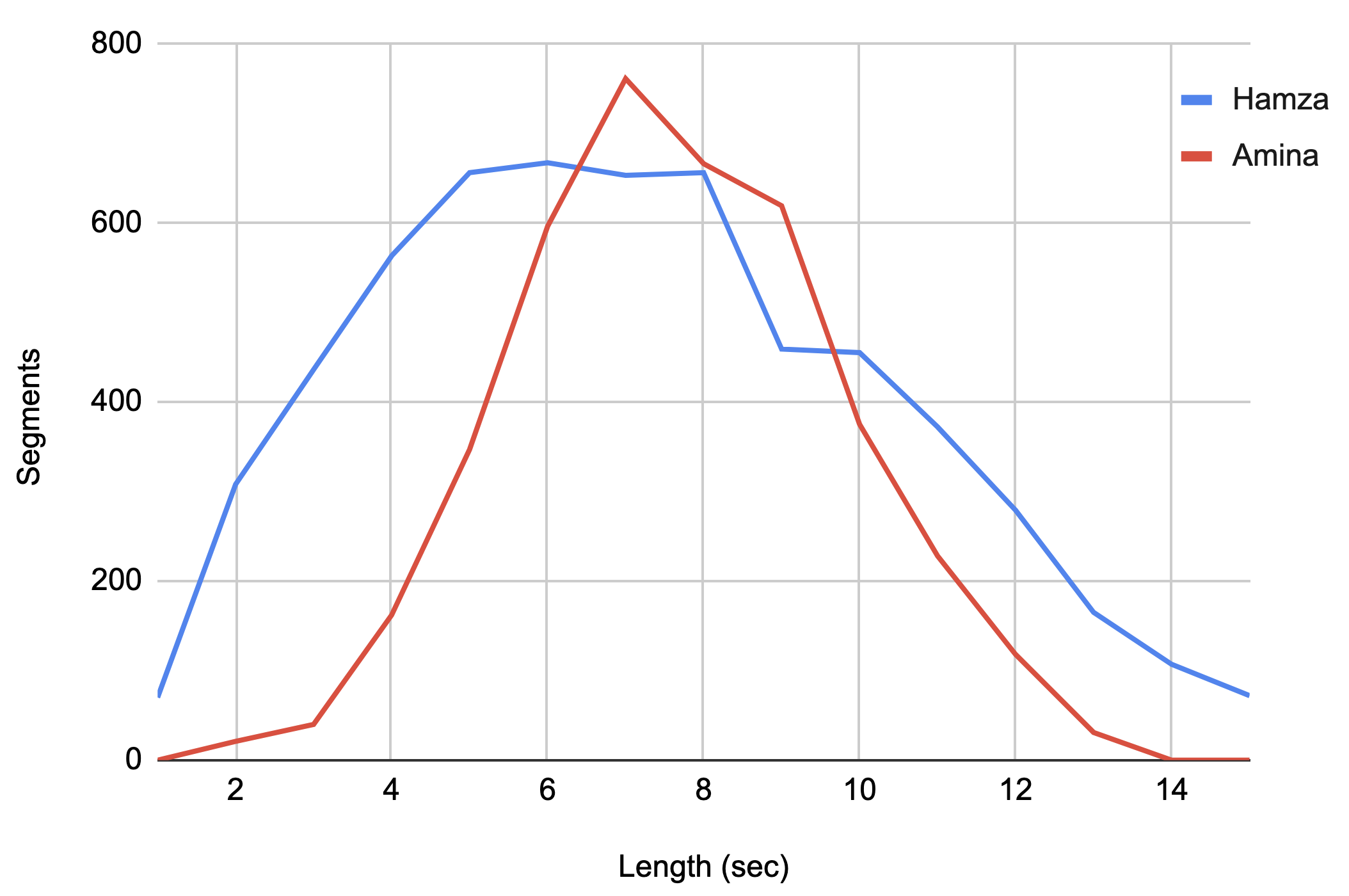}
     \caption{Distribution of segments lengths per speakers}
     \label{fig:datadistribution}
     \vspace{-1mm}
 \end{figure}

We acquired high quality speech data recorded at a sampling rate of 44kHz from two speakers. A female speaker \emph{Amina} 
was recorded reading selected passages  mainly from children books in Modern Standard Arabic. The data contains 3964 segments and 50,714 words in total. The style for this recording is expressive. The second data \emph{Hamza} was recorded by a male speaker and in neutral style. This data contains 6005 segments and 80,409 words in total. Figure~\ref{fig:datadistribution} shows the segments length distribution for each of the speakers. For both of the speakers, the average length of the segments is around 7 seconds or around 12 words per segment.

\subsubsection{Preprocessing}

Data preprocessing steps involve: i) diacritization, ii) speech transcript matching, iii) segmentation, and iv) vowel normalization and silence reduction.

\textbf{Diacritization} Arabic has two types of vowels; namely long vowels, which are explicitly written in the text, and short vowels (aka diacritics) which are typically omitted in modern writings as native speakers can infer them based on contextual information. In order to read Arabic words properly, readers need to restore the missing diacritics and this is important for machines to pronounce the text correctly.
We diacritized the text using \texttt{Farasa} \cite{abdelali-etal-2016-farasa}. 
Although Farasa gives an accuracy above 94\% 
the automatic diacritized data was, neverthless, 
reviewed by a language expert to ensure the accuracy of the 
annotations. This is important as some cases (for example named entities and foreign words) are often even challenging for a native speaker let alone for the automatic system.  
It's worth mentioning that we built a text normalization layer to convert digits, abbreviations, and special symbols to words to be fully diacritized by Farasa. Due to Arabic complexity and ambiguity, this conversion was not trivial in many cases.

\textbf{Speech Transcript Matching} Although native speakers don't require short vowels to correctly pronounce a word, in some rare cases they may make mistake of pronouncing a word with a wrong vowel. Rather than correcting the speaker which might require going back to the studio and re-record the segment again, we 
opted to change the transcript in such cases to reflect what was spoken. This will save both time and efforts required from the speaker and the recording studio.

\textbf{Segmentation} Due to the limitation of neural architectures to handle long audio samples~\cite{shen2018natural}, the data is sampled into frames of 10 seconds in average. The segmentation has to consider the sentence boundaries and not to break nor the context or the prosody. In general cases, long silences between segments is a good indicator but exception were found when related context or supplemental material that is still considered a part of the sentence still comes after a long pause. 

\textbf{Text Normalization} This includes spelling out numbers, fractions, abbreviations and titles into their textual format such as ``16.43'' to ``{\small \<ستة عشر وأربعة  وثلاثين جزء من المئة>}'' (stp Ecr wOrbEp  wvlAvyn jzC mn AlmQp)\footnote{Using Safe Buckwalter Arabic encoding} or ``{\small \<وقال أ. د. ماجد>}'' (wqAl O. d. mAjd)  to ``{\small\<وقال الأستاذ الدكتور ماجد>}'' (wqAl AlOstAV Aldktwr mAjd).

\subsubsection{Models}

We trained three models based on Tacotron-1 \cite{wang2017tacotron}, Tacotron-2 \cite{shen2018natural} and Transformer TTS~\cite{10.1609/aaai.v33i01.33016706}  recipes
. 
The choice of these models was driven mainly by: Real-time decoding and high-quality voice.

\textbf{Model Tacotron1} builds on top of RNN sequence-to-sequence architecture. It includes an encoder, an attention-based decoder, and a post-processing module. The former takes text as characters and generates a mel-spectrogram. The post-processing module then generates waveform from the mel-spectogram. Tacotron1 uses a CBHG-based encoder which consists of a bank of 1-D convolutional filters, followed by highway networks and a bidirectional gated recurrent unit (GRU). The decoder is a content-based tanh attention decoder that generates an 80-band mel-scale spectrogram as the target. Finally we use \textbf{WaveRNN} \cite{kalchbrenner2018efficient} on top to generate waveforms from the generated mel-spectograms. WaveRNN is a single layered RNN network that generates raw audio samples.

\textbf{Model Tacotron2} follows the same recipe as Tacotron1 i.e. RNN-based sequence-to-sequence encoder-decoder architecture, it consists of a bi-directional LSTM-based encoder and a unidirectional LSTM-based decoder with location sensitive attention~\cite{Zhang2018ForwardAttention}. Additionally, the models employs different vocoder to generate waveforms. We used the \textbf{WaveGlow} \cite{prenger2018waveglow}, a flow-based network capable of generating high quality speech from melspectograms. WaveGlow is a generative model that generates audio by sampling from zero mean spherical Gaussian distribution. It uses 12 coupling layers and 12 invertible $1 \times 1$ convolutions.

\textbf{Model ESPnet Transformer TTS} 
Inspired by Neural Machine Translation, Transformer TTS~\cite{10.1609/aaai.v33i01.33016706} adapts multi-head self-attention mechanism and feed forward strategy to build an encoder-decoder model that would convert a sequence of inputs characters into an output sequence of acoustic features (log Mel-filter bank features), the model provide an adventage over the former models in the training speed as it uses a feed forward network compared to recurrent network based-models. Similarly to Tacotron1 and Tacotron2 models, Transformer TTS requires a vocoder to further convert the Mel features into wave form. We used Parallel WaveGAN~\cite{Yamamoto2020ParallelWA}  a non-autoregressive WaveNet that uses generative adversarial network to convert the Mel-filter bank sequences to a waveform. 
\squeezeupless

\section{Evaluation}

To evaluate the performance of each of the models, We built an evaluation test set 
composed of 100 sentences of varying lengths, collected from six domains including: Culture, Economy, Literature, Politics, Sports, and Technology. 
The sentences were collected between Jan 1st to Jan 20th, 2022. They include excerpts from current topics and news. 
We decoded each sentence using the models and for each of the voices. This resulted in a pool of 600 audio files to evaluate. We 
carried automatic and manual (subjective) evaluations described below:

\subsection{Automatic Evaluation}

We used state-of-the-art Arabic ASR system ~\cite{HUSSEIN2022101272} to decode the audio files generated by our TTS models.
The ASR system 
gives state of the art performance on a number of standard data sets such as MGB-3~\cite{MGB38268952} and MGB-5~\cite{MGB59003960}.
We then compare the generated transcripts against the input sentences for which TTS outputs are generated.
As the ASR system generates unvowelized text, we strip short vowels from the reference original text to allow a fair comparison. We used standard evaluation metrics Word Error Rate (WER) and Charecter Error Rate (CER). Table~\ref{asr_eval_results} shows the results using the automatic approach. The system built using ESPnet2 
gave the lowest WER and CER. Additionally, the neutral voice ``Hamza'' achieved a lower error rate 
when compared to the expressive ``Amina''. This highlights the challenges dealing with non-monotonic voices which are typically richer and has more features that the network needs to capture~\cite{Mellotron2019Valle}. For Amina, Tacotron1 results are not worse than the leading ESPnet2 system; which potentially means that Tacotron1 is better at handling richer features. Tacotron2 suffers more from deletion, and substitution errors, this is the main cause for the CER/WER to be higher than other models.


\subsection{Qualitative Evaluation}

We recruited 14 individuals (7 females and 7 males) to carry the manual subjective evaluation. The participants were instructed to listen to the audio and give their opinion on the speech quality using a scale from 1 to 5; The five-category MOS scale~\cite{Guski1997EvaluatingSoundQuality}: 5 = excellent, 4 =
good, 3 = fair, 2 = poor, 1 = bad. Each participants was presented with a set of 15 random samples from the pool. 
The overall results presented in Table~\ref{mos_results} shows that the participants favored ESPNet:Hamza 
and 
Tacotron1:Amina. 
The results of ESPNet:Hamza 
are very comparable to the Tacotron1:Hamza. 
The results also shows that participants preferred the neutral voice over expressive one. Literature also reports that typically evaluators prefer neutral over expressive and expressivity is better perceived when the samples have a high quality~\cite{tahon:hal-01623916}.
The 
qualitative results are closely aligned with automatic evaluation, the differences in CER/WER between ESPNet:Amina and Tacotron1:Amina are less pronounced when compared to Hamza. 
\squeezeupless
\subsection{Speed}
Lastly, another metric to evaluate the system, we used Real-time Factor (RTF): the ratio of the speech generation time to the utterance duration. Such measure is very crucial and essential in the deployment of any system, especially for real-time use. For a system to be considered real-time, RTF should be $<=$ 1~\cite{Pratap2020ScalingASR}. Having a low RTF, will ensure that the system latency is reasonable and acceptable and indicate that the system can be used in real-time applications. Table~\ref{rtf_eval_results} shows the average RTF for the three systems running on a 4 Cores Intel(R) Xeon(R) CPU E5-2640 v4 @ 2.40GHz and 32Gb of RAM and powered by NVIDIA Tesla V100 SXM2 32Gb GPU. The end-to-end ESPnet2 system, is the clear winner with a an RTF equal to 0.09 which is 1.5 and 17 times faster than Tacotron2 and Tacotron1 respectively. None of the systems run real-time on CPU. Our fastest system ESPnet2 runs at a speed of 4.24xRT.

\begin{table}[]
\centering
\begin{tabular}{l|c|c|c|c}
                  & \multicolumn{2}{c|}{Amina} & \multicolumn{2}{c}{Hamza} \\ \cline{2-5} 
                  & CER         & WER         & CER         & WER         \\ \hline
ESPnet2           & \textbf{17.47}  & \textbf{40.42}       & \textbf{8.01}        & \textbf{24.87}       \\ \hline
Tacotron1         & 22.51       & 43.98       & 27.48       & 46.12       \\ \hline
Tacotron2         & 40.76       & 64.80       & 82.38       & 93.62   \\ \hline   
\end{tabular}
 \caption{CER and WER evaluation results.\label{asr_eval_results}}
\end{table}

\begin{table}[]
\centering
\begin{tabular}{l|r|r}
        & \multicolumn{1}{l|}{Amina} & \multicolumn{1}{l}{Hamza} \\ \hline
ESPnet2  & 3.57                      & 4.40                      \\ \hline
Tacotron1        & 4.21                      & 4.38                      \\ \hline
Tacotron2      & 3.49                      & 2.34   \\  \hline                
\end{tabular}
 \caption{MOS evaluation results for the three systems. \label{mos_results}}
\end{table}
\squeezeupless

\begin{table}[htb!]
\centering
\begin{tabular}{l|c|c}
 & \multicolumn{2}{c}{RTF} \\ \hline
Model & GPU & CPU \\ \hline
ESPnet2 & \textbf{0.09} & 4.24\\ \hline
Tacotron1 & 1.66 & - \\  \hline
Tacotron2 & 0.14 & - \\ \hline
\end{tabular}
 \caption{Realtime Factor evaluation results.\label{rtf_eval_results}}
\end{table}

   



\section{Conclusion}
We presented NatiQ Arabic text-to-speech system, a system based on end-to-end framework that combines Transformer encoder-decoder and WaveGAN vocoder. The system was evaluated using subjective metric, Mean Opinion Score and objective Speed, WER and CER. The system achieved a MOS of 4.35 and 4.72 for Amina and Hamza
respectively. Such performance is very comparable to English systems~\cite{wang2017tacotron,shen2018natural} 
.  
For the expressive speaker, the performance of the system still lags behind the neutral one. This is due to the complex and rich features encoded in expressive voice. 
We plan to explore different techniques that exploits the additional features in the voice such as ~\cite{Liu2020ExpressiveTTS} which aim to combine frames and style information as two objective functions to optimize while training the model.         


\bibliography{anthology,custom}
\bibliographystyle{acl_natbib}




\end{document}